\documentclass[preprint,12pt,nonatbib]{elsarticle}

\usepackage{amssymb}
\usepackage{amsmath}
\usepackage[dvipsnames]{xcolor}
\usepackage{subcaption}
\usepackage{placeins}
\usepackage{graphicx}
\usepackage[numbers]{natbib}
\usepackage{hyperref}
\usepackage[automake,acronym]{glossaries}
\usepackage{booktabs}

\journal{Neural Netwoks}

\makeglossaries
\newacronym{afd}{ARD}{Average Random Distance}
\newacronym{ai}{AI}{Artificial Intelligence}
\newacronym{auroc}{AUROC}{Area Under the Receiver Operating Characteristic Curve}
\newacronym{bt}{BT}{Barlow Twin}
\newacronym{cnn}{CNN}{Convolutional Neural Network}
\newacronym{dinov2}{DINOv2}{self-DIstillation with NO labels Version 2}
\newacronym{dino}{DINO}{self-DIstillation with NO labels}
\newacronym{gji}{GJI}{Generalized Jaccard Index}
\newacronym{h0}{H0}{H-optimus-0}
\newacronym{huad}{CAD}{colon adenomas}
\newacronym{infonce}{InfoNCE}{Info Noise Contrastive Estimation}
\newacronym{mocov2}{MoCoV2}{Momentum Contrast Version 2}
\newacronym{mpp}{mpp}{microns per pixel}
\newacronym{swav}{SwAV}{Swapping Assignment between Views}
\newacronym{tcga}{TCGA}{The Cancer Genome Atlas}
\newacronym{vit}{ViT}{Vision Transformer}
\newacronym{wsi}{WSI}{Whole Slide Image}

\begin{document}
\begin{frontmatter}
	\title{Transformation Behavior of Images in Latent Space}
	\affiliation[1]{
		organization={Department of Applied Tumor Biology, Institute of Pathology, Heidelberg University Hospital},
		addressline={Im Neuenheimer Feld 224},
		postcode={69120},
		city={Heidelberg},
		country={Germany}
	}
	\affiliation[2]{
		organization={National Center for Hereditary Tumor Syndromes, University Hospital Bonn},
		addressline={Venusberg-Campus 1},
		postcode={53127},
		city={Bonn},
		country={Germany}
	}
	\affiliation[3]{
		organization={Department of Internal Medicine I, University Hospital Bonn},
		addressline={Venusberg-Campus 1},
		postcode={53127},
		city={Bonn},
		country={Germany}
	}
	\affiliation[4]{
		organization={Leibniz Institut für Wissensmedien},
		addressline={Schleichstraße 6},
		postcode={72076},
		city={Tübingen},
		country={Germany}
	}

	\author[1]{Christian Zöllner}
	\author[1]{Mozzam Motiwala}
	\author[4]{Gerrit Anders}
	\author[1]{Aysel Ahadova}
	\author[2,3]{Robert Hüneburg}
	\author[2,3]{Jacob Nattermann}
	\author[1]{Matthias Kloor}
	\begin{abstract}
		Training of neural networks for histopathology classification tasks typically relies on data encoding into latent space, which reduces complexity and improves performance. There are several encoder networks available, either pretrained on general image datasets such as ImageNET, or specifically on histopathological images. Training of encoder networks should be adapted to downstream tasks, allowing encoding of biologic/diagnostic content while rendering networks invariant to label-irrelevant transformations.

This paper investigates the effect of classical image transformation on the latent space, using networks provided by Lunit Inc. and Bioptimus, both focusing on pathological images, and by Meta Research Team. We assess variance of embeddings resulting from standard data transformations by comparing original and transformed image embeddings and by contrasting them with random, unrelated embeddings, using image tiles from hematoxylin/eosin-stained sections available in a colorectal tissue dataset and the publicly accessible TCGA dataset.

Our findings show that embeddings of original and transformed images are closer to each other than to random embeddings, indicating robustness to transformations. However, they are not fully invariant, revealing that the encoder networks do not completely neutralize transformation effects in latent space, explaining why transformation-mediated augmentation of datasets can improve performance. Significant differences were observed between general and histopathology-specific encoder networks.
	\end{abstract}
	\begin{keyword}
		embedder \sep data transformation \sep unsupervised learning \sep machine learning \sep latent space

	\end{keyword}
\end{frontmatter}

\section{Introduction}
\label{sec:introduction}

\gls{ai} is playing a transformative role in modern medicine, particularly through its ability to analyze and interpret complex medical data.
Machine learning algorithms, especially neural networks,
have brought significant advancements in disease detection, classification, and personalized treatment plans.
However, the scarcity of labeled data poses a major challenge.
In the medical domain, collecting large, annotated datasets requires the involvement of skilled professionals,
making it a time-consuming and expensive process.
This remains a key obstacle to maximizing the potential of \gls{ai} in healthcare.

Common strategies to mitigate data scarcity include data transformations, which artificially generate additional data points with valid information, and embedder networks, which enhance the usability of available data so that less is needed.
Data transformations artificially enlarge datasets
by creating modified versions of the data that retain the same information for the neural network to learn from.
For example, given a dataset $D$,
consisting of training samples $S_i$ and corresponding labels $L_i$,
a transformation $T_j$ is applied to the sample $S_i$
resulting in $S'_i$ which still matches the original label $L_i$\cite{mumuniDataAugmentationComprehensive2022}.

\begin{equation}
    T(S_i, L_i) \rightarrow (S'_i, L_i)
    \label{eq:data_transformation}
\end{equation}

For image-based datasets, transformations include techniques like mirroring, cropping, flipping,
and color augmentations\cite{shortenSurveyImageData2019, mumuniDataAugmentationComprehensive2022}.

In recent years, another approach has emerged to reduce the time- and cost-intensive effort of labeling:
so-called embedder networks are trained on an abundance of unlabeled data using unsupervised learning algorithms.
These networks map data to a latent space to create compact yet informative representations,
which are subsequently used by classification or regression networks.

\begin{equation}
    C(E(S_i)) = C(X_i) = L_i
    \label{eq:embedder_classification_network}
\end{equation}

Here, the latent space vector $X_i$ is an abstract representation of the data, typically living in $\mathbb{R}^n$.
Key properties of effective latent spaces,
as noted in the literature\cite{bengioRepresentationLearningReview2014, schmidhuberLearningFactorialCodes1992, ridgewaySurveyInductiveBiases2016, achilleEmergenceInvarianceDisentanglement2018},
are:

\begin{enumerate}
    \item Smoothness: Small variations to samples should lead to small shifts in latent space.
    In other words, samples close to each other in data space should be close to each other in latent space.
    \item Dimensionality:
    The complexity or dimensionality of the latent space should be lower than the space of the original data,
    lowering computational costs.
    \item Multi-Task-Relevance:
    While removing redundant and unimportant information from the sample,
    the representation in the latent space should still include enough features to be of use in multiple downstream tasks.
    \item Disentanglement: Each dimension should represent a statistically independent feature.
\end{enumerate}

To get a better understanding of embedder networks, we explore four widely used unsupervised training methods:
\newline

\noindent\textbf{\gls{bt}}:
This training method makes use of the concept that not relevant information should be excluded from the latent space.
Therefore, it creates two copies from the input data,
which are then passed individually through two randomly selected combinations of transformations.
The altered images are passed through the network, outputting a latent vector.
The correlation matrix of these latent vectors is then used to calculate the loss,
optimizing the network
to produce the same results for any combination of the used transformations\cite{zbontarBarlowTwinsSelfSupervised2021}. \newline

\noindent\textbf{\gls{mocov2}}:
While the \gls{bt} method only provides positive samples to cluster similar images in latent space,
the \gls{mocov2}\cite{chenImprovedBaselinesMomentum2020} training also includes negative samples.
During the training process, an image is sampled and put through an embedder.
An augmented version of the image is put through a second embedder network of the same architecture but different weights.
It is called a momentum embedder
because it will be updated indirectly by interpolating the new weights between the first embedder and the old weights.
The latent vector of the momentum encoder is put into a queue, beside embeddings of previous steps,
replacing the oldest one.
Then the dot product is calculated between the latent vector of the current image and each latent vector in the queue,
which they call keys in analogy to dictionary keys.
This gives a set of logits, allowing to calculate a possibility for each key that it belongs to the image.
Using an \gls{infonce} loss the network learns to select the correct matching key
and lowers the chance to select a key belonging to a negative sample. \newline

\noindent\textbf{\gls{swav}}:
Like the other two methods, \gls{swav}\cite{caronUnsupervisedLearningVisual2021a}
also makes heavy use of image augmentations.
It starts with one image and sample $N$ randomly selected transformations, creating $N$ augmentations.
These are then passed through the network, yielding $N$ latent vectors belonging to the same class.
For the next step, we need so-called cluster vectors.
These are predefined and learnable vectors in the latent space.
So they have the same dimensionality as the latent vectors of the image.
The paper mentioned that they got the best results when choosing a magnitude more cluster vectors than final classification labels.
The core idea is now to match each latent vector to a cluster vector.
This is like assigning each augmentation of the image to an abstract feature cluster,
then shuffling these and trying to predict the so-called codes,
which are calculated each iteration using the Sinkhorn Knopp algorithm\cite{cuturiSinkhornDistancesLightspeed2013}.
These codes serve as artificial class labels,
crafted so that the network does not collapse to none of the trivial solutions,
either assigning each image only to one cluster or assigning them the total equally distributed.
In the end, the network yields the most frequently occurring abstract features of the dataset as cluster vectors,
and can order incoming new data according to these clusters. \newline

\noindent\textbf{\gls{dino}}:
The last presented training method \gls{dino}\cite{caronEmergingPropertiesSelfsupervised2021} is similar to the \gls{mocov2} approach.
It also uses image transformations to create different views of an image,
before passing them through two embedder networks, here called student and teacher network.
Like in \gls{mocov2},
they have the same architecture but different weights
and the teacher embedder is trained with a moving average from the student network.
The key difference here is that the student network gets smaller images (local views) than the teacher network
(global views), which is the same as a cropping transformation.
Additionally, the output of the teacher is centralized along the mean of the batch, before computing the loss.
The student network is now trained to reproduce the same latent vectors as the teacher network.
The principle of the training can be summarized
as teaching a network to produce the same latent vector from many small image patches as from a few global ones. \newline

Each of these methods utilizes data transformations to train embedder networks, with the shared goal of mapping data with equal or similar relevant information to nearby, or identical locations in the latent space. 
This leads to the assumption that a perfect embedder network, trained over infinite time on infinite data, would discard all information altered by these transformations, resulting in an invariant output.
But since resources are finite, no such perfect network exists.
This raises a pivotal question: How do embedder networks behave under commonly applied data transformations?
Investigating this question by analyzing representative networks forms the central focus of this paper.
While rooted in a theoretical inquiry, this investigation may also yield practical benefits: Measuring invariance could provide a direct and interpretable way to assess the quality of latent spaces, enabling systematic comparisons between different embedder networks without depending exclusively on downstream task performance.
\section{Related Work}
\label{sec:related_work}

Many papers in the literature highlight the importance of proper data preprocessing, improving performance and avoiding overfitting by using data transformations\cite{liFuzzybasedDataTransformation2011, shortenSurveyImageData2019, temizEffectsDataEnrichment2022, shiReducingPredictionError2000}.
Also, nearly all modern classifiers make use of embedder networks to improve their performance.
Some also use not only visual but also text-based data in a shared latent space\cite{zhouPathM3MultimodalMultitask2024, zhengBenchmarkingPathCLIPPathology2024, huangVisuallanguageFoundationModel2023}.
However, little work has been conducted on the quality of latent spaces.

But we could not find any paper investigating if the relationship between using data transformations in the preprocessing pipeline and embedder networks relying on the invariance of these transformations.
Specifically, to the best of our knowledge, no systematic analysis has been performed so far to directly investigate the impact of data transformation in the preprocessing pipeline on variation of the resulting latent space vectors using distinct embedder networks.
\section{Experiments}\label{sec:experiments}
The central research question of this paper is how latent vectors, created by a network, which has been trained in an unsupervised manner, behave under transformations used during its training.
To answer this, we pass a set of images and their corresponding augmented images through the network and measure the L2-Norm distance between the original and transformed embedding vector.

We will also analyze the disentanglement of latent features as a property of efficient latent spaces.
Therefore, we look at the dimensional distribution of the differences, comparing which dimensions change most under different transformations.
Below we will give a more detailed look at the dataset, networks and transformation we will use, as well as the exact experiment setup.

\begin{figure}
    \centering
    \includegraphics[width=\textwidth]{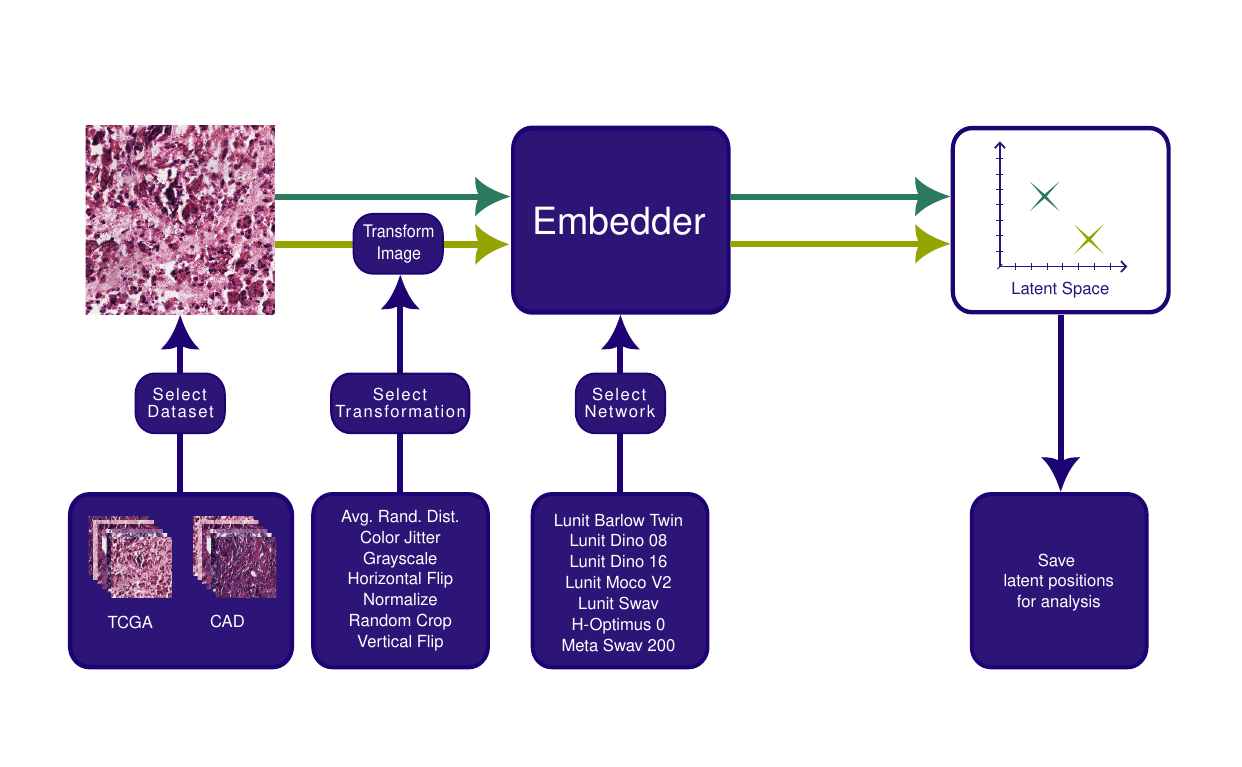}
    \caption{Each experiment is done by first selecting a dataset and resolution, than a transformation and lastly an embedder network. The images of the dataset are passed through the embedder and a copy of the images are first transformed and then passed through the embedder. The resulting latent vectors are stored for further processing.}
    \label{fig:experiment_flow_chart}
\end{figure}

\subsection{Datasets}\label{subsec:datasets}
We conducted our experiments on two distinct~\gls{wsi} datasets, both displaying tissue from colon tumors.
The first dataset is the open-source Colon Adenocarcinoma collection from~\gls{tcga}\cite{grossmanSharedVisionCancer2016}, comprising 1330~\gls{wsi} with a total size of 430.91~GB\@.
To address generalizability of our findings, we also included a non-public image data set of \gls{huad} available at the Department of Applied Tumor Biology, Heidelberg University Hospital, and the University Hospital Bonn, consisting of 520~\gls{wsi}, summing up to 234~GB in total.
We processed the slides into tiles of 256 by 256 pixels and used the Canny Edge Detection algorithm\cite{cannyComputationalApproachEdge1986} to automatically detect and exclude background tiles.
The networks we tested are trained on~\gls{wsi} with resolutions of 0.25~\gls{mpp} and 0.5~\gls{mpp}.
Consequently, we generated tiles at resolutions of 0.25~\gls{mpp} and 0.5~\gls{mpp}.
Additionally, to examine any potential alterations in network performance, we also created tiles at 1~\gls{mpp} and 2~\gls{mpp}.
This approach produced eight diverse datasets for our experiments.

\subsection{Networks}\label{subsec:networks}
The networks used in this paper are built upon two different architectures.
The first and maybe most common is the ResNet50\cite{heDeepResidualLearning2016a}, belonging to the class of~\gls{cnn}.
Its core idea is to propagate feature maps through the network with skip connections, adding them to the output of the last residual block, before feeding the results into the next.
The second architecture belongs to the same class as the GPT networks, transformers.
It is called~\gls{vit} and was introduced in the paper ``An image is worth 16x16 words: transformers for image recognition at scale''\cite{dosovitskiyImageWorth16x162021}.
While to this point transformer architectures were mostly used to process natural language, the approach performed as well as the other state-of-the-art~\gls{cnn} of the time.
The images are deconstructed into patches of fixed size, then linearized and padded with a positional encoding.
After that, they are fed into the regular transformer blocks, where they can be processed in parallel at a big scale. \newline

We conducted our analysis using seven different sets of network weights.
As a baseline, we selected a model using the ResNet50 architecture, trained over 200 epochs on the ImageNet dataset\cite{deng2009imagenet} following the~\gls{swav} methodology.
The weights for this model were sourced from Meta Research\cite{FacebookresearchSwav2024}, who provided these along with their publication, ``Unsupervised Learning of Visual Features by Contrasting Cluster Assignments''\cite{caronUnsupervisedLearningVisual2021a}.
Although Meta Research also released models trained for 400 and 800 epochs and other architectural variations, we opted for the 200-epoch ResNet50 weights to ensure comparability with models trained on pathology~\gls{wsi} for the equivalent number of epochs.
We examined five networks used in the study ``Benchmarking Self-Supervised Learning on Diverse Pathology Datasets''\cite{kangBenchmarkingSelfSupervisedLearning2023} by Lunit Inc.
This study provided two networks based on the~\gls{vit} architecture\cite{dosovitskiyImageWorth16x162021} and three on the ResNet50 architecture\cite{heDeepResidualLearning2016a}.
In particular, one of the ResNet50 networks was also trained using the~\gls{swav} method, which allows a direct comparison with the baseline model from Meta Research.
The other two ResNet50 networks use the \gls{mocov2}\cite{chenImprovedBaselinesMomentum2020}~and~\gls{bt}\cite{zbontarBarlowTwinsSelfSupervised2021} trainings methods, while the~\gls{vit} networks both use the~\gls{dino}\cite{caronEmergingPropertiesSelfsupervised2021} trainings method.
The last and newest network we included is the \gls{h0} developed by the Bioptimus company~\cite{hoptimus0}. 
The architecture is also a \gls{vit}, but with 1.1 billion parameter the by far largest network in our study. 
It was trained it on 500,000 slides with a resolution of 0.5~\gls{mpp} with an improved version of the \gls{dino} trainings method \gls{dinov2}~\cite{oquab2024dinov2learningrobustvisual}.

\subsection{Transformations}\label{subsec:transformations}

Each experiment setup selects one of the following transformations: horizontal and vertical flip, random crop, color jitter, conversion to greyscale, color normalization, and random comparison image.
We perform on all tiles a center crop to 224 by 224 pixels, before applying the transformation of interest. 
For the random crop transformation we perform no center crop, to get a consistent image size. 
Since there are many versions of color normalization, we used the method and parameter described in\cite{kangBenchmarkingSelfSupervisedLearning2023}.
The last transformation `random comparison image' acts as a baseline in our experiments.
Since the distances in latent spaces have no unit and are unique for each latent space, the `random comparison image' transformation samples a random, not transformed image from our dataset.
Using this approach, we calculated the average distance and its variation between our samples in the latent space.
By normalizing the measured distances to this average distance, we can compare the different latent spaces to each other.
Through normalizing each measured distance with the average distance to a random embedding in the same latent space, we can now directly compare distances in different latent spaces. 

\subsection{Methodology}\label{subsec:methodology}
In total, 392 different experiments were performed, comprising eight datasets, seven networks, and seven transformations.
Each of them was executed as follows.
A fixed number of tiles was sampled from the dataset and a transformed copy of each tile was created.  
By using the same amount of tiles, independent of the dataset, we ensured that the results were not biased by the dataset size.
Both the original and transformed tiles were passed through the selected network, yielding two latent vectors, which were stored for the following analysis.

The first analysis computed the L2-Norm between these two vectors and normalized these distances with respect to the average distance between the measured latent vectors, followed by calculation of the average and variance.
In the second analysis, we calculated the average difference in each dimension. 
For the last evaluation we again computed the differences per dimension but this time truncated all but the 10 most changing dimensions, comparing the remaining dimensions with other experiments of the same network. 
To quantize the overlap between the different sets of most affected dimensions we used a \gls{gji} \ref{eq:generalized-jaccard-index}. 
Because the Jaccard Index\cite{JaccardIndex} $J$ (\ref{eq:jaccard_index}) itself can only compare two sets at a time we could not use it here.
\begin{equation}
    J(A,B) = \frac{\left| A \cap B \right|}{\left| A \cup B \right|}
    \label{eq:jaccard_index}
\end{equation}
A natural generalization to $n$ sets $A_1,\dots,A_n$ would tend to zero very quickly as $n$ grows and would also not capture partial overlap of the sets. 
\begin{equation}
    J(A_1,\dots,A_n) = \frac
    {\left| \bigcap^{n}_{i=1} A_i \right|}
    {\left| \bigcup^{n}_{i=1} A_i \right|}
\end{equation}
To address this issue, one can define for $k \in \left\{ 2,\dots,n\right\}$ the set $S_k$ and its corresponding similarity $J_k$
\begin{align}
    S_k &= \left\{ x: \text{$x$ belongs to at least k of the $A_i$} \right\} \\
    J_k &= \frac
    {\left| S_k\right|}
    {\left| \bigcup^{n}_{i=1} A_i \right|}
\end{align}
This gives us a way to measure the similarity for pairwise, three-way, and so on overlap. 
To construct now a similarity including all these overlaps we compute the average $M$ of these quantities.
\begin{equation}
   M = \frac{1}{n - 1} \sum^{n}_{m=2} J_m
   \label{eq:generalized-jaccard-index}
\end{equation}
$M$ from (\ref{eq:generalized-jaccard-index}) is now the \gls{gji} we used.

\section{Results}\label{sec:results}

\subsection{L2-Norm Analysis}\label{subsec:l2-norm-analysis}

The average distances between the original and transformed embeddings yielded from the 0.25~\gls{mpp} \gls{tcga} and~\gls{huad} datasets are presented in Fig~\ref{fig:results-avg-free-distance}.
The complete results can be found in the appendix. 
No significant differences with regard to resolution or between the two datasets could be observed. 
The mean distance between an embedding and a randomly chosen image, referred to as~\gls{afd}, was greater than the average distances observed for other transformations.
This column also exhibits the highest variance, being of the same magnitude as the mean itself.
Among transformations, spatial transformations such as flipping and cropping had a smaller impact on the embeddings compared to color-related transformations.
Although differences were observed between the used networks, no strong outliers were found.
For color-related transformations, the highest distance between original and transformed embeddings were detected for Meta's~\gls{swav} and the~\gls{h0} network from Bioptimus, when normalized to~\gls{afd}.
We also see that Meta's~\gls{swav} display consistent high distances for flip transformations, compared to other networks, especially for the vertical. 

\begin{figure}[hbt!]
    \centering
    \begin{subfigure}[b]{1.\textwidth}
        \centering
        \includegraphics[width=0.8\linewidth]{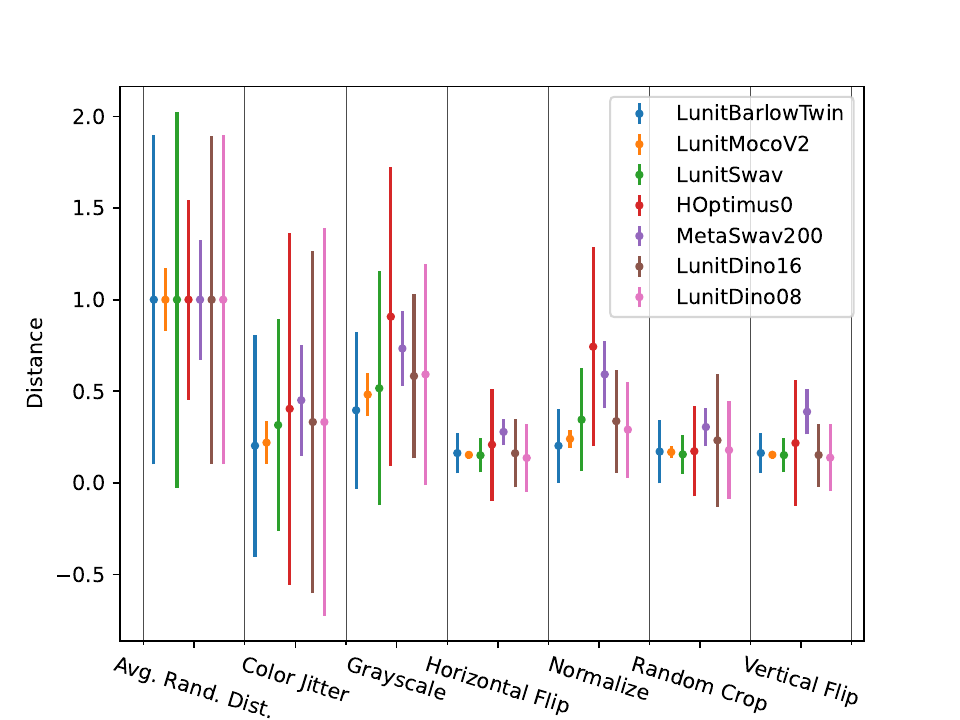}
        \caption{\gls{tcga} at 0.25 \gls{mpp}}
    \end{subfigure}
    \hfill
    \begin{subfigure}[b]{1.\textwidth}
        \centering
        \includegraphics[width=0.8\linewidth]{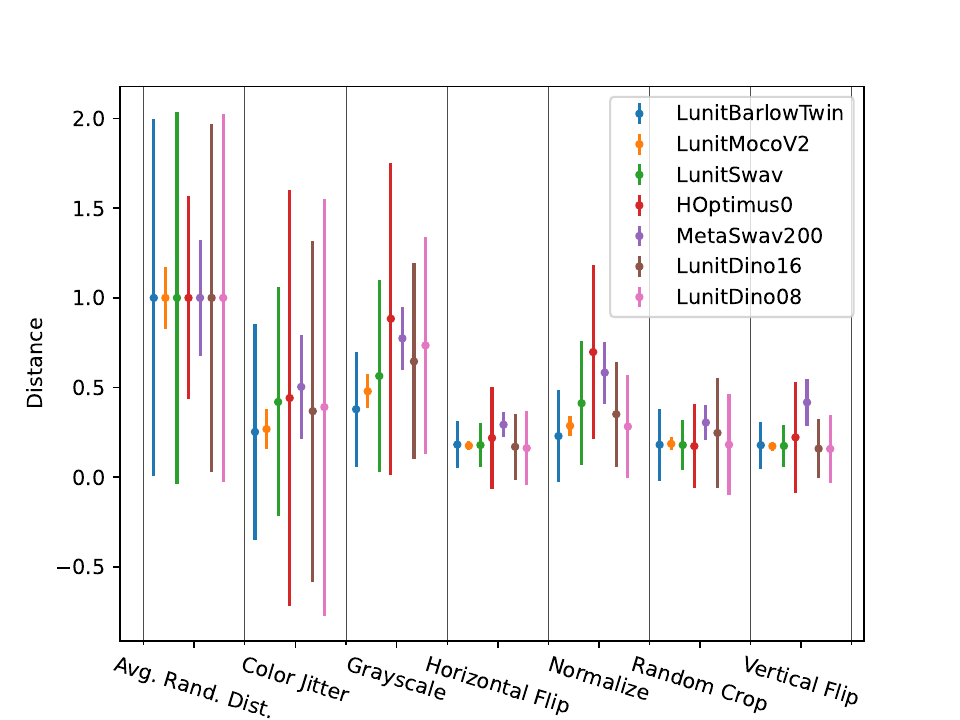}
        \caption{\gls{huad} at 0.25 \gls{mpp}}
    \end{subfigure}
    \caption{Average distances between original and transformed embeddings, normalized to the average free distance between embeddings from the \gls{tcga} and \gls{huad} dataset, on 0.25~\gls{mpp}.}
    \label{fig:results-avg-free-distance}
\end{figure}

\clearpage

\subsection{Dimension Distribution}\label{subsec:dimension-distribution}

As said in the introduction, a well-trained embedder network should filter and disentangle data into well-defined abstract features.
Therefore, one would expect that the effect of a transformation is limited to one or a few dimensions. 

\begin{figure}[h]
    \centering
    \begin{subfigure}[b]{0.49\textwidth}
        \centering
        \includegraphics[width=\textwidth]{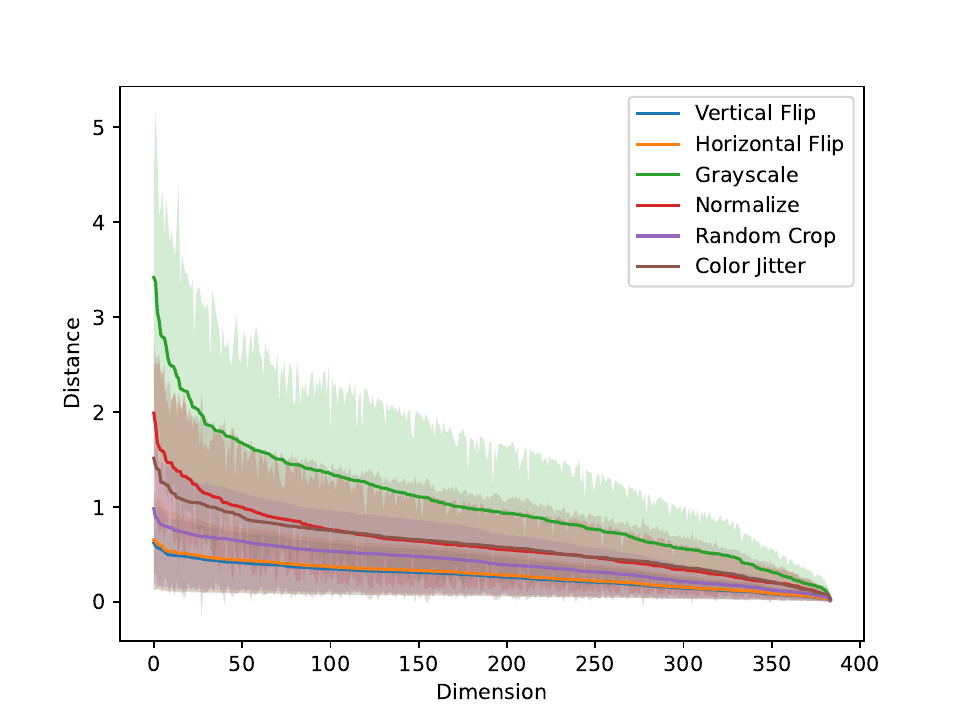}
        \caption{Lunit Dino 16 \gls{tcga} 0.25 \gls{mpp}}
    \end{subfigure}
    \hfill
    \begin{subfigure}[b]{0.49\textwidth}
        \centering
        \includegraphics[width=\textwidth]{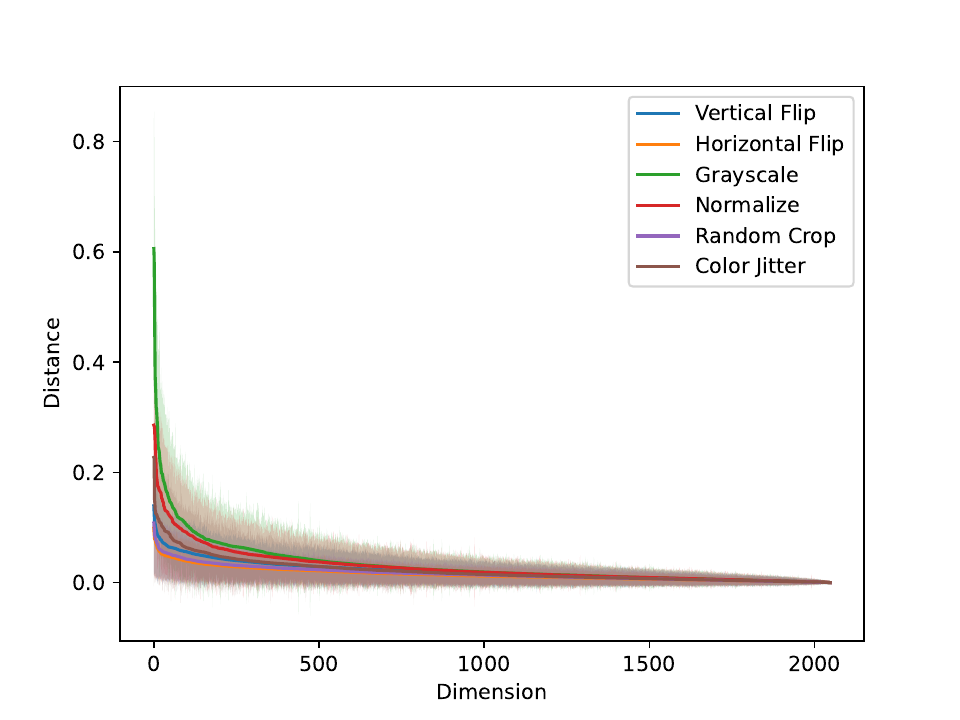}
        \caption{Meta \gls{swav} \gls{tcga} 0.25 \gls{mpp}}
    \end{subfigure}
    \vfill
    \begin{subfigure}[b]{0.49\textwidth}
        \centering
        \includegraphics[width=\textwidth]{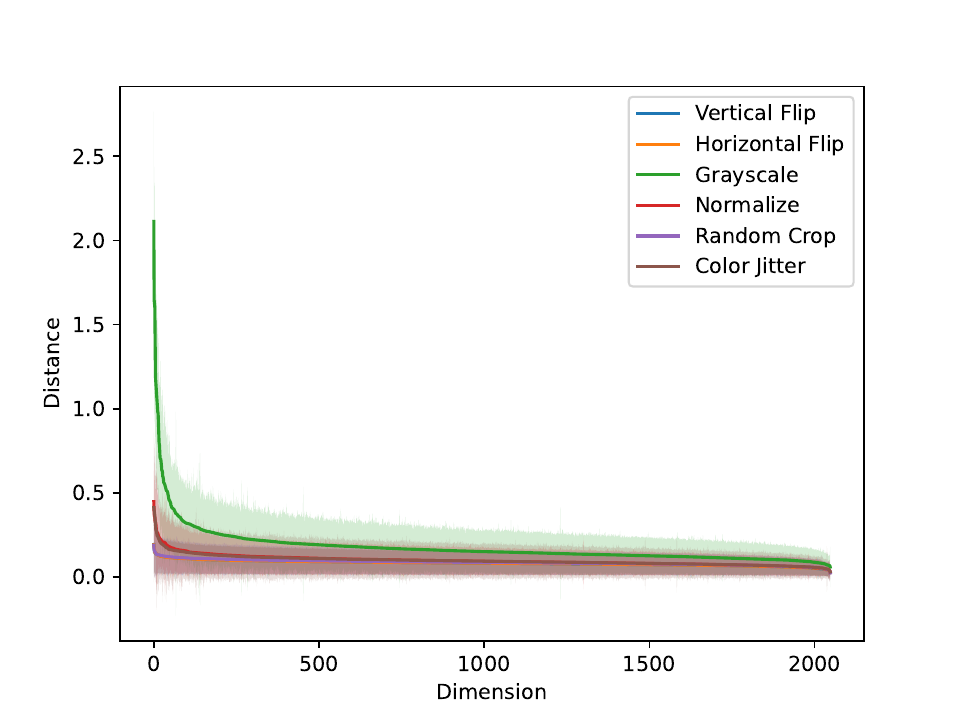}
        \caption{Lunit \gls{bt} \gls{tcga} 2.00 \gls{mpp}}
    \end{subfigure}
    \hfill
    \begin{subfigure}[b]{0.49\textwidth}
        \centering
        \includegraphics[width=\textwidth]{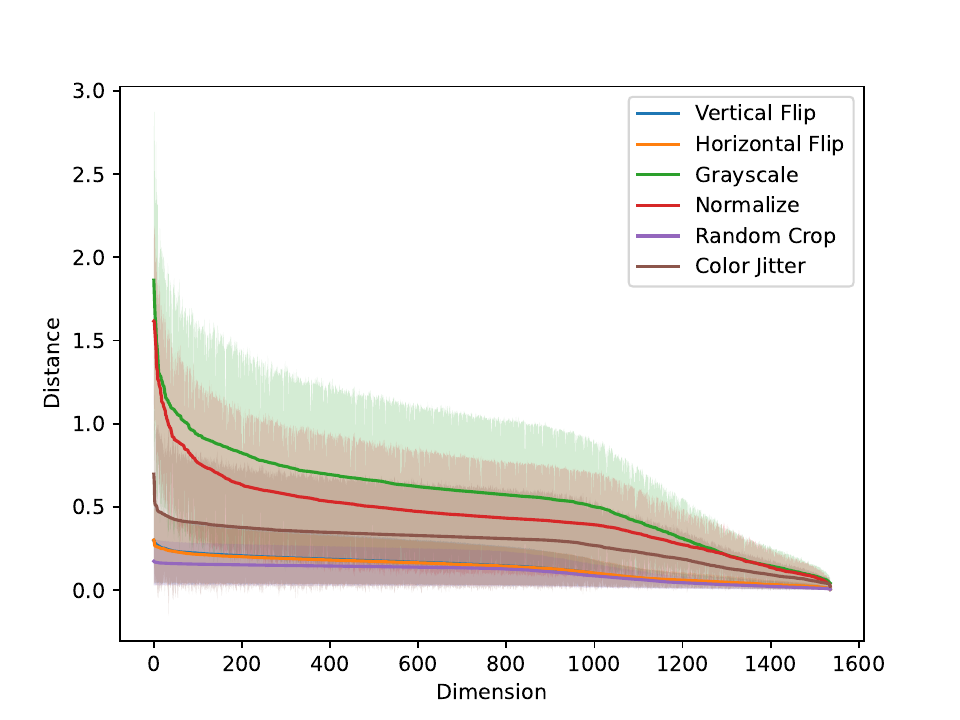}
        \caption{Bioptimus~\gls{h0} \gls{tcga} 0.25 \gls{mpp}}
    \end{subfigure}
    \caption{
        Comparing a subset of the sorted dimensional distance distribution.
        While the overall form stays the same for all networks and datasets, there are differences in the steepness of the distributions.~}
    \label{fig:sorted-histogram-of-distance-per-dimension}
\end{figure}

Figure~\ref{fig:sorted-histogram-of-distance-per-dimension} shows the change per dimensions between embeddings, sorted from the most to the least changing dimension.
Overall, for all networks, datasets, and transformations few dimensions contributed most to the change.
But we can also see that no dimension is entirely invariant under any transformation.
Lunit's~\gls{dino} and Bioptimus~\gls{h0} networks showed a very gentle slope, with many dimensions contributing to each transformation.
The Meta~\gls{swav} network is here comparable to the remaining network from Lunit, showing a steep slope with better feature disentanglement.

\FloatBarrier

\subsection{Dimensional Index Analysis}\label{subsec:dimensional-index-analysis}

\begin{figure}[!ht]
    \centering
    \begin{subfigure}[b]{0.49\textwidth}
        \centering
        \includegraphics[width=\textwidth]{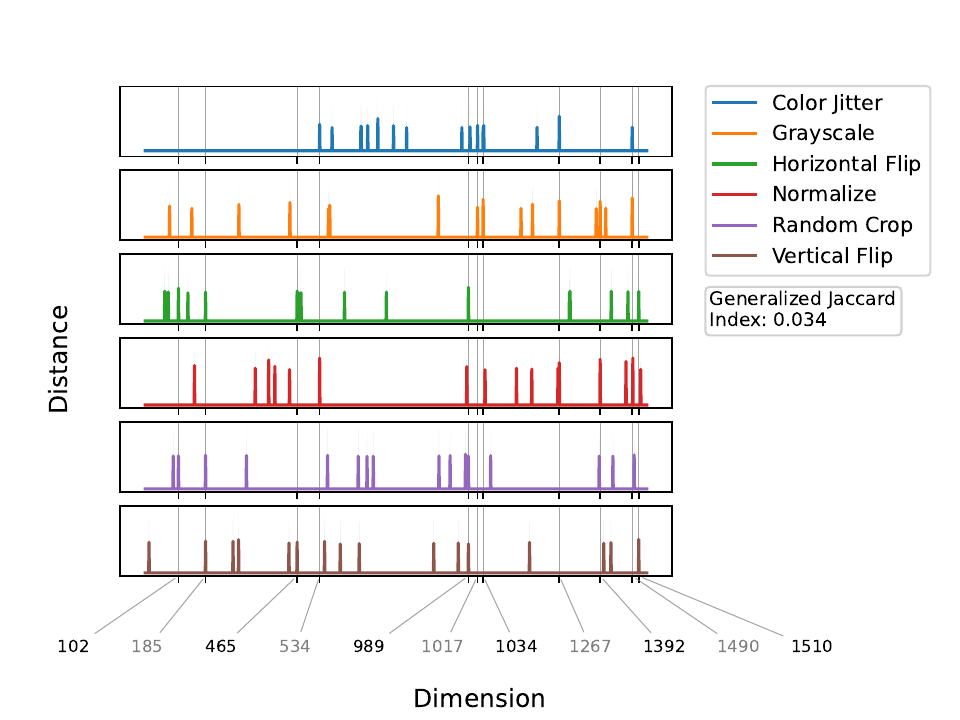}
        \caption{The combination with the smallest overlap was the network from \gls{h0} on the \gls{huad} 0.25 \gls{mpp} dataset.}
        \label{fig:smallest-gji}
    \end{subfigure}
    \hfill
    \begin{subfigure}[b]{0.49\textwidth}
        \centering
        \includegraphics[width=\textwidth]{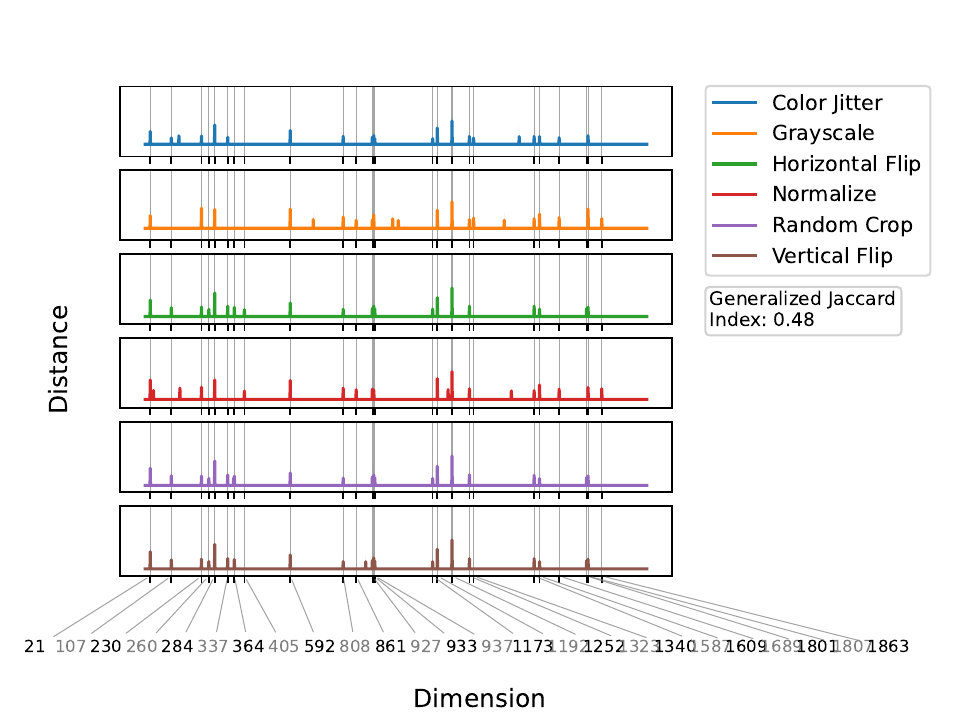}
        \caption{The highest overlap showed the \gls{swav} network from Lunit on the \gls{huad} 2.00 \gls{mpp} dataset.}
        \label{fig:largest-gji}
    \end{subfigure}
    \caption{}
    \label{fig:smallest-and-largest-gji}
\end{figure}

Figure \ref{fig:smallest-and-largest-gji} shows the ten most contributing dimensions for each of the transformations. 
We choosed here to show the network and dataset combination with the smallest (\ref{fig:smallest-gji}) and largest overlap (\ref{fig:largest-gji}).

\begin{table}
\caption{Mean \gls{gji} per network across all datasets.}
\label{tab:mean_gji_per_network}
\begin{tabular}{ll}
\toprule
 & GJI \\
Model &  \\
\midrule
HOptimus0 ViT & 0.091 ± 0.012 \\
LunitBarlowTwin ResNet50 & 0.267 ± 0.026 \\
LunitDino08 ViT & 0.137 ± 0.012 \\
LunitDino16 ViT & 0.178 ± 0.013 \\
LunitMocoV2 ResNet50 & 0.380 ± 0.051 \\
LunitSwav ResNet50 & 0.456 ± 0.051 \\
MetaSwav200 ResNet50 & 0.392 ± 0.017 \\
\bottomrule
\end{tabular}
\end{table}

Also table \ref{tab:mean_gji_per_network} displays the mean \gls{gji} across datasets. 
Relative to their variances the values are quite consistant across the datasets. 
The \gls{h0} network shows by far the smallest overlap, indicating better feature disentanglement.
\FloatBarrier

\section{Discussion}\label{sec:discussion}
In the present study, we examined the effects of histopathology image transformation on latent space embeddings.
We observed that embeddings in latent space were not invariant under any of the transformations and embedder networks tested.
However, quantification of variance revealed, that the distance between two randomly selected image embeddings, on average, was larger than for any other transformation on all networks and datasets.
We also see that the variance of these is of the same magnitude as the distance itself, indicating that the cloud of embeddings is not distributed across the entire embedding space, but rather localized.
The fact that the findings were highly similar for both, the \gls{tcga} and \gls{huad} datasets, suggests that our observations are of general validity for histopathology image datasets.

Notably, we observed that color-based transformations had a greater impact than spatial-based transformation on the embedding shifts across all networks and datasets.
Since different~\gls{wsi} scanners tend to have slight variations in color, it is plausible to assume that scanner types may have significant effects on embeddings in latent space, and change of scanners could cause large shifts.

This complicates classification tasks and underscores the importance of color normalization for improving the generalizability of AI-based classifiers in histopathology.

Zheng et al. \cite{ZHENG2019107} developed a color normalization method, adaptive color deconvolution, which increased the \gls{auroc} from 0.842 to 0.914.
Similarly, Vicory et al. \cite{vicoryAppearanceNormalizationHistology2015} reported an improvement from 0.967 to 0.981 in \gls{auroc} and from 0.899 to 0.962 in accuracy after applying color normalization techniques.

In contrast, the comparatively small effect of spatial transformations on the position of embeddings in latent space indicates that such networks largely disregard this information—as expected—since spatial orientation does not contain label-relevant information in histopathology sections.

We also see that the networks, fine-tuned to pathological images, are more invariant than the baseline network from Meta, trained on ImageNet.
Especially when looking at the difference between `horizontal-' and `vertical flip' transformation, we can see that the network trained on ImageNet undergoes a greater shift for the `vertical flip' compared to the `horizontal flip' transformation.
This is consistent with the used datasets, since the vertical orientation on natural images, on the contrary to pathological images, carries meaning.
We also see that the Meta network has greater displacement for 'horizontal flip' than the other networks, because, again, in natural images it carries meaning if something is on the right or left of an image, but to a lesser extent than the vertical orientation.
We observed that for conceptually distinct transformations, the most strongly affected latent dimensions were often the same, while the shift induced by each transformation was not localized but distributed across all dimensions.
This pattern suggests that the networks does not properly disentangle underlying image features, violating a core principle of a well-structured latent space.

Our findings indicate that while these embedder networks are not flawless, it is unclear if the mappings in latent space through transformations are correlated with worse performance of the networks in downstream task.
A notable example is the H0 network from Bioptimus.
It undergoes comparable large shifts in latent space, but has the best feature disentanglement of all tested networks.
A potential issue would arise if the transformations caused data points with different labels to become more closely clustered, as this would complicate classification.
However, as long as distinct labels occupy separate regions within the latent space, training classification networks on these representations remains possible.

\section{Conclusion and Outlook}\label{sec:conclusion-and-outlook}

For each network and dataset, we find that embeddings shift in latent space when the source images undergo classical image transformation.
The L2 distance between the original and transformed embeddings is small, particularly for spatial transformations, compared to the average distance between embeddings in latent space, but it remains clearly non-zero.
Moreover, we observe that the same latent dimensions are frequently affected by multiple transformations, indicating incomplete disentanglement of features within the latent space.
One could argue that these findings are unsurprising, as no network can be trained with infinite resources, which as in theory could yield perfect representations.
However, the experiments presented a valuable and intuitive tool for comparing the latent spaces of different networks.
Our results underscore the continued importance of classical image augmentations in preprocessing pipelines, especially given the observed deviations from ideal transformation-invariant embeddings.
Whether high sensitivity to transformations negatively impacts downstream task performance remains an open question.
Investigating this correlation will be the subject of future work.
Ultimately, the pursuit of training embedder networks yielding suitable sized and structured latent spaces remains a central challenge in representation learning.
Further research is needed to deepen our understanding of latent space structure and improve the disentanglement of learned features.
\newpage

\FloatBarrier
\printglossary[type=\acronymtype]

\clearpage
\section{Declaration of generative AI}\label{sec:declaration-of-generative-ai}

During the preparation of this work, the author used openai-gpt-4o to improve readability and language of the work.
After using this tool, the author reviewed and edited the content as needed and take full responsibility for the content of the published article.

\section{Ethics Statement}\label{sec:ethics-statement}
The usage of the tissue samples in this study was conducted in accordance with the Declaration of Helsinki and approved by the ethics commission of the medical faculty of the university clinic Heidelberg (approval number: S-348/2023).
Written informed consent was obtained from all participants for the use of their tissue samples in scientific research.
Data from the \gls{tcga} were used under their open-access data use policy and contain fully anonymized human tissue images.
No new human or animal experiments were performed by the authors.

\section{Acknowledgements}

The excellent technical support provided by Nina Nelius, Ricarda Mehr, Jonathan Doerre is gratefully acknowledged.
Parts of the work have been supported by grants from Deutsche Krebshilfe (German Cancer Aid) and Klaus Tschira Foundation.

The authors acknowledge support by the state of Baden-Württemberg through bwHPC
and the German Research Foundation (DFG) through grant INST 35/1597-1 FUGG.

\bibliographystyle{elsarticle-num}
\bibliography{bib}

@article{mumuniDataAugmentationComprehensive2022,
    title = {Data Augmentation: A Comprehensive Survey of Modern Approaches},
    shorttitle = {Data Augmentation},
    author = {Mumuni, Alhassan and Mumuni, Fuseini},
    date = {2022-12-01},
    year = {2022},
    journal = {Array},
    shortjournal = {Array},
    volume = {16},
    pages = {100258},
    issn = {2590-0056},
    doi = {10.1016/j.array.2022.100258},
    url = {https://www.sciencedirect.com/science/article/pii/S2590005622000911},
    urldate = {2025-01-13},
}

@article{shortenSurveyImageData2019,
    title = {A Survey on Image Data Augmentation for Deep Learning},
    author = {Shorten, Connor and Khoshgoftaar, Taghi M.},
    date = {2019-07-06},
    year = {2019},
    journal = {Journal of Big Data},
    shortjournal = {Journal of Big Data},
    volume = {6},
    number = {1},
    pages = {60},
    issn = {2196-1115},
    doi = {10.1186/s40537-019-0197-0},
    url = {https://doi.org/10.1186/s40537-019-0197-0},
    urldate = {2025-01-13},
}

@article{bengioRepresentationLearningReview2014,
  author={Bengio, Yoshua and Courville, Aaron and Vincent, Pascal},
  journal={IEEE Transactions on Pattern Analysis and Machine Intelligence}, 
  title={Representation Learning: A Review and New Perspectives}, 
  year={2013},
  volume={35},
  number={8},
  pages={1798-1828},
  doi={10.1109/TPAMI.2013.50},
}

@inproceedings{deng2009imagenet,
    author = {Deng, Jia and Dong, Wei and Socher, Richard and Li, Li-Jia and Li, Kai and Fei-Fei, Li},
    booktitle = {Computer Vision and Pattern Recognition, 2009. CVPR 2009. IEEE Conference on},
    organization = {IEEE},
    pages = {248--255},
    title = {Imagenet: A large-scale hierarchical image database},
    url = {https://ieeexplore.ieee.org/abstract/document/5206848/},
    year = 2009
}

@misc{zhouPathM3MultimodalMultitask2024,
    title = {{PathM}3: a multimodal multi-task multiple instance learning framework for whole slide image classification and captioning},
    url = {http://arxiv.org/abs/2403.08967},
    doi = {10.48550/arXiv.2403.08967},
    shorttitle = {{PathM}3},
    number = {arXiv:2403.08967},
    publisher = {arXiv},
    author = {Zhou, Qifeng and Zhong, Wenliang and Guo, Yuzhi and Xiao, Michael and Ma, Hehuan and Huang, Junzhou},
    urldate = {2024-08-20},
    date = {2024-07-23},
    year = {2024},
    eprinttype = {arxiv},
    eprint = {2403.08967 [cs]},
}

@misc{kangBenchmarkingSelfSupervisedLearning2023,
    title = {Benchmarking self-supervised learning on diverse pathology datasets},
    url = {http://arxiv.org/abs/2212.04690},
    doi = {10.48550/arXiv.2212.04690},
    number = {{arXiv}:2212.04690},
    publisher = {{arXiv}},
    author = {Kang, Mingu and Song, Heon and Park, Seonwook and Yoo, Donggeun and Pereira, Sérgio},
    urldate = {2024-04-02},
    date = {2023-04-18},
    year = {2023},
    eprinttype = {arxiv},
    eprint = {2212.04690 [cs]},
}

@misc{dosovitskiyImageWorth16x162021,
    title = {An image is worth 16x16 words: Transformers for image recognition at scale},
    url = {http://arxiv.org/abs/2010.11929},
    doi = {10.48550/arXiv.2010.11929},
    shorttitle = {An image is worth 16x16 words},
    number = {{arXiv}:2010.11929},
    publisher = {{arXiv}},
    author = {Dosovitskiy, Alexey and Beyer, Lucas and Kolesnikov, Alexander and Weissenborn, Dirk and Zhai, Xiaohua and Unterthiner, Thomas and Dehghani, Mostafa and Minderer, Matthias and Heigold, Georg and Gelly, Sylvain and Uszkoreit, Jakob and Houlsby, Neil},
    urldate = {2024-11-05},
    date = {2021-06-03},
    year = {2021},
    eprinttype = {arxiv},
    eprint = {2010.11929},
}

@article{zhengBenchmarkingPathCLIPPathology2024,
    title = {Benchmarking {PathCLIP} for pathology image analysis},
    issn = {2948-2933},
    url = {https://doi.org/10.1007/s10278-024-01128-4},
    doi = {10.1007/s10278-024-01128-4},
    journal = {Journal of Imaging Informatics in Medicine},
    shortjournal = {J Digit Imaging. Inform. med.},
    author = {Zheng, Sunyi and Cui, Xiaonan and Sun, Yuxuan and Li, Jingxiong and Li, Honglin and Zhang, Yunlong and Chen, Pingyi and Jing, Xueping and Ye, Zhaoxiang and Yang, Lin},
    urldate = {2024-11-05},
    date = {2024-07-09},
    year = {2014},
    langid = {english},
}

@article{huangVisuallanguageFoundationModel2023,
    title = {A visual-language foundation model for pathology image analysis using medical twitter},
    volume = {29},
    issn = {1546-170X},
    doi = {10.1038/s41591-023-02504-3},
    pages = {2307--2316},
    number = {9},
    journal = {Nature Medicine},
    shortjournal = {Nat Med},
    author = {Huang, Zhi and Bianchi, Federico and Yuksekgonul, Mert and Montine, Thomas J. and Zou, James},
    date = {2023-09},
    year = {2023},
    pmid = {37592105},
}

@misc{zbontarBarlowTwinsSelfSupervised2021,
    title = {Barlow twins: Self-supervised learning via redundancy reduction},
    url = {http://arxiv.org/abs/2103.03230},
    doi = {10.48550/arXiv.2103.03230},
    shorttitle = {Barlow twins},
    number = {{arXiv}:2103.03230},
    publisher = {{arXiv}},
    author = {Zbontar, Jure and Jing, Li and Misra, Ishan and {LeCun}, Yann and Deny, Stéphane},
    urldate = {2024-11-18},
    date = {2021-06-14},
    year = {2021},
    eprinttype = {arxiv},
    eprint = {2103.03230},
}

@article{grossmanSharedVisionCancer2016,
    title = {Toward a shared vision for cancer genomic data},
    volume = {375},
    issn = {1533-4406},
    doi = {10.1056/NEJMp1607591},
    pages = {1109--1112},
    number = {12},
    journal = {The New England Journal of Medicine},
    shortjournal = {N Engl J Med},
    author = {Grossman, Robert L. and Heath, Allison P. and Ferretti, Vincent and Varmus, Harold E. and Lowy, Douglas R. and Kibbe, Warren A. and Staudt, Louis M.},
    date = {2016-09-22},
    year = {2016},
    pmid = {27653561},
    pmcid = {PMC6309165},
}

@article{cannyComputationalApproachEdge1986,
    title = {A Computational Approach to Edge Detection},
    volume = {{PAMI}-8},
    issn = {1939-3539},
    url = {https://ieeexplore.ieee.org/document/4767851},
    doi = {10.1109/TPAMI.1986.4767851},
    pages = {679--698},
    number = {6},
    journal = {{IEEE} Transactions on Pattern Analysis and Machine Intelligence},
    author = {Canny, John},
    urldate = {2024-11-18},
    date = {1986-11},
    year = {1989},
    note = {Conference Name: {IEEE} Transactions on Pattern Analysis and Machine Intelligence},
}

@inproceedings{heDeepResidualLearning2016a,
  author={He, Kaiming and Zhang, Xiangyu and Ren, Shaoqing and Sun, Jian},
  booktitle={2016 IEEE Conference on Computer Vision and Pattern Recognition (CVPR)}, 
  title={Deep Residual Learning for Image Recognition}, 
  year={2016},
  volume={},
  number={},
  pages={770-778},
  doi={10.1109/CVPR.2016.90}
}

@inproceedings{cuturiSinkhornDistancesLightspeed2013,
 author = {Cuturi, Marco},
 booktitle = {Advances in Neural Information Processing Systems},
 editor = {C.J. Burges and L. Bottou and M. Welling and Z. Ghahramani and K.Q. Weinberger},
 publisher = {Curran Associates, Inc.},
 title = {Sinkhorn Distances: Lightspeed Computation of Optimal Transport},
 url = {https://proceedings.neurips.cc/paper_files/paper/2013/file/af21d0c97db2e27e13572cbf59eb343d-Paper.pdf},
 volume = {26},
 year = {2013}
}

@inproceedings{caronEmergingPropertiesSelfsupervised2021,
    title = {Emerging Properties in Self-Supervised Vision Transformers},
    booktitle = {2021 {{IEEE}}/{{CVF International Conference}} on {{Computer Vision}} ({{ICCV}})},
    author = {Caron, Mathilde and Touvron, Hugo and Misra, Ishan and Jegou, Herve and Mairal, Julien and Bojanowski, Piotr and Joulin, Armand},
    date = {2021-10},
    year = {2021},
    pages = {9630--9640},
    publisher = {IEEE},
    location = {Montreal, QC, Canada},
    doi = {10.1109/ICCV48922.2021.00951},
    url = {https://ieeexplore.ieee.org/document/9709990/},
    urldate = {2024-12-06},
    eventtitle = {2021 {{IEEE}}/{{CVF International Conference}} on {{Computer Vision}} ({{ICCV}})},
    isbn = {978-1-6654-2812-5},
    langid = {english}
}

@misc{chenImprovedBaselinesMomentum2020,
    title = {Improved baselines with momentum contrastive learning},
    url = {http://arxiv.org/abs/2003.04297},
    doi = {10.48550/arXiv.2003.04297},
    number = {{arXiv}:2003.04297},
    publisher = {{arXiv}},
    author = {Chen, Xinlei and Fan, Haoqi and Girshick, Ross and He, Kaiming},
    urldate = {2024-12-06},
    date = {2020-03-09},
    year = {2020},
    eprinttype = {arxiv},
    eprint = {2003.04297 [cs]},
}

@misc{caronUnsupervisedLearningVisual2021a,
    title = {Unsupervised learning of visual features by contrasting cluster assignments},
    url = {http://arxiv.org/abs/2006.09882},
    doi = {10.48550/arXiv.2006.09882},
    number = {{arXiv}:2006.09882},
    publisher = {{arXiv}},
    author = {Caron, Mathilde and Misra, Ishan and Mairal, Julien and Goyal, Priya and Bojanowski, Piotr and Joulin, Armand},
    urldate = {2024-12-04},
    date = {2021-01-08},
    year = {2021},
    eprinttype = {arxiv},
    eprint = {2006.09882 [cs]},
}

@inproceedings{FacebookresearchSwav2024,
  title={Unsupervised Learning of Visual Features by Contrasting Cluster Assignments},
  author={Caron, Mathilde and Misra, Ishan and Mairal, Julien and Goyal, Priya and Bojanowski, Piotr and Joulin, Armand},
  booktitle={Proceedings of Advances in Neural Information Processing Systems (NeurIPS)},
  year={2020}
}

@article{vicoryAppearanceNormalizationHistology2015,
    title = {Appearance normalization of histology slides},
    volume = {43},
    issn = {0895-6111},
    url = {https://www.sciencedirect.com/science/article/pii/S0895611115000658},
    doi = {10.1016/j.compmedimag.2015.03.005},
    pages = {89--98},
    journal = {Computerized Medical Imaging and Graphics},
    shortjournal = {Computerized Medical Imaging and Graphics},
    author = {Vicory, Jared and Couture, Heather D. and Thomas, Nancy E. and Borland, David and Marron, J. S. and Woosley, John and Niethammer, Marc},
    urldate = {2024-12-04},
    date = {2015-07-01},
    year = {2015},
}

@article{schmidhuberLearningFactorialCodes1992,
    title = {Learning Factorial Codes by Predictability Minimization},
    author = {Schmidhuber, Jürgen},
    date = {1992-11-01},
    year = {1992},
    journal = {Neural Computation},
    shortjournal = {Neural Computation},
    volume = {4},
    number = {6},
    pages = {863--879},
    issn = {0899-7667},
    doi = {10.1162/neco.1992.4.6.863},
    url = {https://doi.org/10.1162/neco.1992.4.6.863},
    urldate = {2025-01-14}
}

@misc{ridgewaySurveyInductiveBiases2016,
      title={A Survey of Inductive Biases for Factorial Representation-Learning}, 
      author={Karl Ridgeway},
      year={2016},
      eprint={1612.05299},
      archivePrefix={arXiv},
      primaryClass={cs.LG},
      url={https://arxiv.org/abs/1612.05299}, 
}

@inproceedings{achilleEmergenceInvarianceDisentanglement2018,
  author={Achille, Alessandro and Soatto, Stefano},
  booktitle={2018 Information Theory and Applications Workshop (ITA)}, 
  title={Emergence of Invariance and Disentanglement in Deep Representations}, 
  year={2018},
  volume={},
  number={},
  pages={1-9},
  doi={10.1109/ITA.2018.8503149},
}

@article{temizEffectsDataEnrichment2022,
    title = {Effects of Data Enrichment with Image Transformations on the Performance of Deep Networks},
    volume = {2},
    issn = {2822-2296},
    url = {http://arxiv.org/abs/2306.07724},
    doi = {10.56038/ejrnd.v2i2.23},
    pages = {23--33},
    number = {2},
    journal = {The European Journal of Research and Development},
    shortjournal = {{EJRnD}},
    author = {Temiz, Hakan},
    urldate = {2025-01-15},
    date = {2022-06-07},
    year = {2022},
    eprinttype = {arxiv},
    eprint = {2306.07724 [cs]},
}

@article{shiReducingPredictionError2000,
    title = {Reducing prediction error by transforming input data for neural networks},
    volume = {14},
    issn = {0887-3801},
    url = {https://ascelibrary.org/doi/10.1061/%28ASCE%290887-3801%282000%2914%3A2%28109%29},
    doi = {10.1061/(ASCE)0887-3801(2000)14:2(109)},
    pages = {109--116},
    number = {2},
    journal = {Journal of Computing in Civil Engineering},
    author = {Shi, Jonathan Jingsheng},
    urldate = {2025-01-15},
    date = {2000-04-01},
    year = {2000},
    note = {Publisher: American Society of Civil Engineers},
}

@article{liFuzzybasedDataTransformation2011,
    title = {A fuzzy-based data transformation for feature extraction to increase classification performance with small medical data sets},
    volume = {52},
    issn = {0933-3657},
    url = {https://www.sciencedirect.com/science/article/pii/S0933365711000182},
    doi = {10.1016/j.artmed.2011.02.001},
    pages = {45--52},
    number = {1},
    journal = {Artificial Intelligence in Medicine},
    shortjournal = {Artificial Intelligence in Medicine},
    author = {Li, Der-Chiang and Liu, Chiao-Wen and Hu, Susan C.},
    urldate = {2025-01-15},
    date = {2011-05-01},
    year = {2011},
}

@misc{oquab2024dinov2learningrobustvisual,
      title={DINOv2: Learning Robust Visual Features without Supervision}, 
      author={Maxime Oquab and Timothée Darcet and Théo Moutakanni and Huy Vo and Marc Szafraniec and Vasil Khalidov and Pierre Fernandez and Daniel Haziza and Francisco Massa and Alaaeldin El-Nouby and Mahmoud Assran and Nicolas Ballas and Wojciech Galuba and Russell Howes and Po-Yao Huang and Shang-Wen Li and Ishan Misra and Michael Rabbat and Vasu Sharma and Gabriel Synnaeve and Hu Xu and Hervé Jegou and Julien Mairal and Patrick Labatut and Armand Joulin and Piotr Bojanowski},
      year={2024},
      eprint={2304.07193},
      archivePrefix={arXiv},
      primaryClass={cs.CV},
      url={https://arxiv.org/abs/2304.07193}, 
}

@misc{hoptimus0,
  author = {Saillard, Charlie and Jenatton, Rodolphe and Llinares-López, Felipe and Mariet, Zelda and Cahané, David and Durand, Eric and Vert, Jean-Philippe},
  title = {H-optimus-0},
  url = {https://github.com/bioptimus/releases/tree/main/models/h-optimus/v0},
  year = {2024},
}

@article{ZHENG2019107,
title = {Adaptive color deconvolution for histological WSI normalization},
journal = {Computer Methods and Programs in Biomedicine},
volume = {170},
pages = {107-120},
year = {2019},
issn = {0169-2607},
doi = {https://doi.org/10.1016/j.cmpb.2019.01.008},
url = {https://www.sciencedirect.com/science/article/pii/S0169260718312161},
author = {Yushan Zheng and Zhiguo Jiang and Haopeng Zhang and Fengying Xie and Jun Shi and Chenghai Xue}
}

@article{JaccardIndex,
    ISSN = {0028646X, 14698137},
    URL = {http://www.jstor.org/stable/2427226},
    author = {Paul Jaccard},
    journal = {The New Phytologist},
    number = {2},
    pages = {37--50},
    publisher = {[Wiley, New Phytologist Trust]},
    title = {The Distribution of the Flora in the Alpine Zone},
    urldate = {2025-08-29},
    volume = {11},
    year = {1912}
}

\end{document}